\documentclass[review,3p]{elsarticle}

\usepackage{lineno,hyperref}
\modulolinenumbers[1]

\usepackage{amssymb}
\usepackage{subfigure}
\usepackage{algorithm}
\usepackage{algpseudocode}

\journal{undetermined}

\begin{document}
\begin{frontmatter}                           

%
\title{IEOPF: An Active Contour Model for Image Segmentation with Inhomogeneities Estimated by Orthogonal Primary Functions}

\author[mysecondaryaddress]{Chaolu Feng\corref{mycorrespondingauthor}}
\cortext[mycorrespondingauthor]{Corresponding author}
\ead{fengchaolu@cse.neu.edu.cn}


\address[mysecondaryaddress]{School of Computer Science and Engineering,
Northeastern University, Shenyang, Liaoning 110014, China}


\begin{abstract}
Image segmentation is still an open problem especially when intensities of the interested objects are overlapped due to the presence of intensity inhomogeneity (also known as bias field). To segment images with intensity inhomogeneities, a bias correction embedded level set model is proposed where Inhomogeneities are Estimated by Orthogonal Primary Functions (IEOPF). In the proposed model, the smoothly varying bias is estimated by a linear combination of a given set of orthogonal primary functions. An inhomogeneous intensity clustering energy is then defined and membership functions of the clusters described by the level set function are introduced to rewrite the energy as a data term of the proposed model. Similar to popular level set methods, a regularization term and an arc length term are also included to regularize and smooth the level set function, respectively. The proposed model is then extended to multichannel and multiphase patterns to segment colourful images and images with multiple objects, respectively. It has been extensively tested on both synthetic and real images that are widely used in the literature and public BrainWeb and IBSR datasets. Experimental results and comparison with state-of-the-art methods demonstrate that advantages of the proposed model in terms of bias correction and segmentation accuracy.
\end{abstract}


\end{frontmatter}


\section{Introduction}
\label{sec:intro}
\begin{figure}[!h]
\centering
{\includegraphics[width=0.75\textwidth]{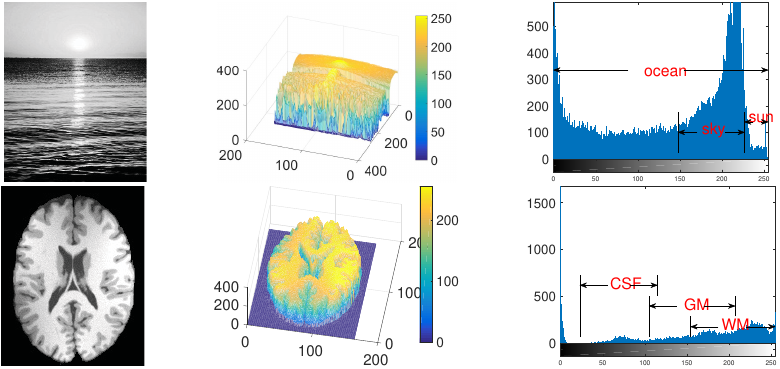}}
\caption{Intensity overlaps of interested objects in a camera image (upper) and an MR brain image (lower). \label{fig:inhomogeneities}}
\end{figure}

Image segmentation is a fundamental but one of the most important problems in pattern recognition and computer vision \citep{feng2016image}. In general, it aims at separating an image into several parts corresponding to meaningful objects. Inner elements (i.e., pixels for 2D images or voxels for 3D images) of each part, recognized as components of a desired object, are considered as having an identical characteristic in terms of shape, structure, or texture \citep{paragios2002geodesic}. As well known, image segmentation has been extensively studied for decades and many efforts have been devoted to proposing effective methods, but it is still a challenging task to extract interested objects accurately from a complex image \citep{feng2015segmentation, feng2016segmentation}. In particular, if the image is corrupted by noise and bias field, intensity homogeneity of the image will be destroyed due to intensity overlaps between different objects, which certainly brings challenges to classical segmentation methods that are based upon edge detection or thresholding. Unfortunately, intensity inhomogeneities exist in most of real-world images inevitably. Fig.~\ref{fig:inhomogeneities} gives an example to demonstrate negative effects of inhomogeneities on intensity distribution of a camera captured nature image and a medical brain image.

As mentioned earlier, a variety of segmentation methods have been proposed in the literature where active contour models (ACMs) have been extensively studied as one class of the most popular ones and have proven to be specially effective for image segmentation due to their ability to elastically deform and delineate object boundaries with smooth and closed contours in sub-pixel accuracy \citep{paragios2000geodesic, he2008comparative}. The fundamental idea of ACMs is to introduce a contour to represent boundaries of interested objects and then drive the contour moving toward its interior normal direction under some constraints. The constraints are generally contained in a predefined energy function and the function will finally get its minimal value when the contour stops on true boundaries of the desired objects. However, there are inherent drawbacks of traditional ACMs, e.g., initialization sensitivity and difficulties associated with topological changes in merging and splitting of the evolving contour. Therefore, since the active contour model was proposed by Kass {\emph{et al.}} in \citep{kass1988snakes}, many efforts have been devoted to developing improved methods to overcome the inherent drawbacks \citep{menet1990active,osher2001level}. As one of the most important improvements of ACMs, level set methods regard the active contour as the zero level set contour of a predefined one-dimension-higher function \citep{osher2003geometric}. Motion of the contour is implied in evolution of the entire level set function under a principled energy minimization framework instead of directly driving the contour itself. Therefore, interesting elastic behaviours of the active contour are preserved with topological changes of the contour efficiently handled by the evolution of the level set function. In addition, level set methods are easily extended to a higher dimension and prior knowledge of interested objects can be incorporated into their energy framework to guide the zero level set contour moves close to the desired boundaries \citep{paragios2003level,feng2016simultaneous}.

\begin{figure}[!h]
\centering
{\includegraphics[width=0.75\textwidth]{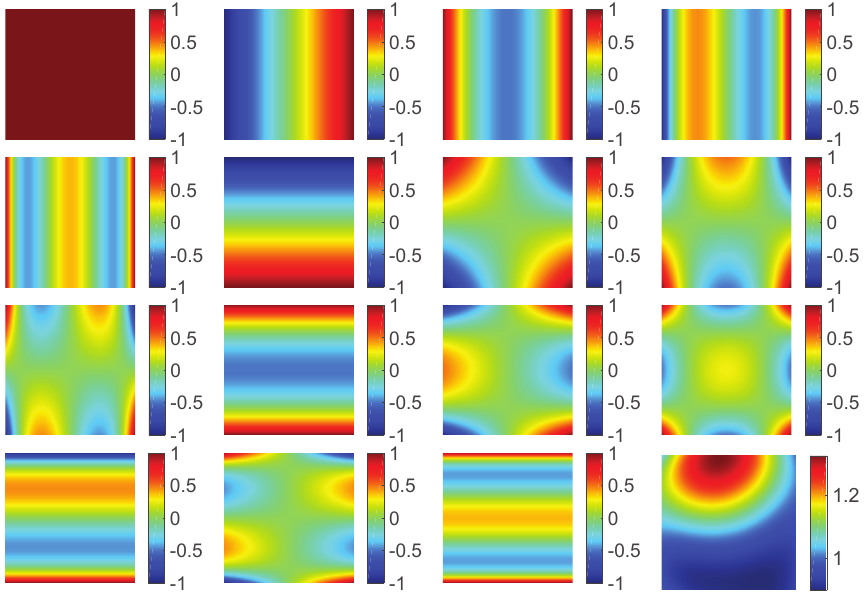}}
\caption{15 2D orthogonal Legendre functions and the bias field (down-right) estimated using these functions with weighting coefficients 1.05, -0.05, -0.06, 0.01, 0.01, -0.20, 0.04, 0.12, -0.02, 0.02, 0.01, -0.02, 0.05, -0.03, -0.01 from up-left to down-right, respecitvely. \label{fig:biasfuncs}}
\end{figure}

Existing level set methods are usually classified into edged-based and region-based methods depending on whether an edge indicator or a region descriptor is used to guide the motion of the zero level set contour. Edged-based level set methods are particularly efficient to recognize boundaries with sharp gradient, but they are not only generally sensitive to noise, but also often suffer from the boundary leakage problem especially in the vicinity of objects with weak boundaries \citep{li2010distance}. The drawbacks are overcome in region-based level set methods by introducing region descriptors based on statistical information of the image in general to identify each region of interest \citep{zhang2010active}. In this paper, a region based level set model is proposed where bias correction is embedded in the model. Specifically, inhomogeneous intensities in the model are estimated by orthogonal primary functions. A demonstration of orthogonal Legendre functions in fitting smooth two dimensional functions is given in Fig.~\ref{fig:biasfuncs}. We further extend the proposed model to segment multichannel images and images with multiple objects.

The rest of this paper is organized as follows. We first briefly review related work and some typical ACM models in Section \ref{sec_relatedworks}. Details of the proposed model IEOPF are presented in Section \ref{sec_pf}. Experimental results of the proposed model on synthetic and natural images that are widely used in the literature and comparison with state-of-the-art models on BrainWeb and IBSR image repositories are given in Section \ref{sec_expresult}. We analyse and discuss relationship and improvement of the proposed model with state-of-the-art model, its robustness to initialization, and coefficient impact in Section \ref{sec_discuss}. This paper is finally summarized in Section \ref{sec_con_future}.

\section{Related Work}
\label{sec_relatedworks}
Let $\Omega\subset R^n$ be a n-dimensional continuous domain and $I$ be an image defined on the domain. That is to say, the observed image can be viewed as a mapping from $\Omega$ to $R$. In general, the problem of image segmentation using ACMs is in fact to find a optimal contour $C$ to separate the image $I$ into two non-overlapping parts, i.e $\Omega_1$ and $\Omega_2$, each of which is regarded as a desired object. {{Note that as a great diversity of level set methods have been proposed in the literature \citep{chan2001active,vese2002multiphase,li2008minimization,li2010distance,li2011level,feng2016image_LINC}, we take widely used symbols in this paper to avoid confusion.}}

\subsection{Classical Mumford-Shah functional model}
To find the optimal contour $C$, Mumford and Shah proposed an energy based segmentation model via an energy functional in \citep{mumford1989optimal}. The basic idea of this classical model is to find a pair of $(u, C)$ for a given image $I$, where $u$ is a nearly piecewise smooth approximation of $I$. The general form of this functional can be written as
\begin{equation}
\label{eq_ms_model}
E_{MS}(u,C)=\int (u-I)^2 d{\bf x} + \mu\int_{\Omega\setminus C}\mid\triangledown u\mid d{\bf x} + \nu|C| d{\bf x}
\end{equation}
where $\mu$ and $\nu$ are positive weighting coefficients. Note that unless otherwise specified, integrations are all performed on the entire image domain $\Omega$ in this paper. 

{\bf{Remark 1.}} When the contour $C$ is exactly located on the desired boundaries and $u$ is piecewise smooth enough to approximate $I$, this functional takes its minimal value and vice versa. However, it is not easy to find the optimal solution of above defined energy functional due to different natures of the unknown $C$ and $u$ and the non-convexity of the functional as well.

\subsection{Chan-Vese's piecewise constant model}
To overcome the difficulties in solving Eq.~(\ref{eq_ms_model}), Chan and Vese proposed a piecewise constant case of the Mumford-Shah model in \citep{chan2001active}, which have proven to be particularly influential in binary segmentation. In the well known CV model, the contour $C$ that separates the image $I$ into two parts is considered as the $0$-level set contour of a level set function $\phi$, i.e. $C \triangleq \lbrace {\bf x}: \phi({\bf x})=0 \rbrace$. Function values of $\phi$ are opposite in sign on either side of the $0$-level set contour. We let the level set function $\phi$ take respectively negative and positive values in regions $\Omega_1$ and $\Omega_2$ which locate inside and outside the $0$-level set contour $C$, i.e. $\Omega_1 \triangleq \lbrace {\bf x}: \phi({\bf x})<0\rbrace$ and $\Omega_2 \triangleq \lbrace {\bf x}: \phi({\bf x})>0\rbrace$. Thus, membership functions $M_1(\phi({\bf x}))=1-H(\phi({\bf x}))$ and $M_2(\phi({\bf x}))=H(\phi({\bf x}))$ can be respectively used to represent these two regions by making $M_1(\phi({\bf x}))=1$ for ${\bf x} \in \Omega_1$, $M_2(\phi({\bf x}))=1$ for ${\bf x} \in \Omega_2$, and otherwise both of them are $0$. Note that $H$ is the Heaviside function. Then, the energy functional of the CV model is defined by
\begin{eqnarray}
\label{eq_CV_dataterm}
E_{CV}(c_1,c_2,\phi) &=& \sum_{i=1}^2\int\mid I({\bf x})-c_i\mid^2 M_i(\phi({\bf x}))d{\bf x} \nonumber \\
 &+&\mu {\cal A}(\phi)+\nu {\cal L}(\phi)
\end{eqnarray}
where ${\cal A}(\phi)=\int (1-H(\phi({\bf x})))d{\bf x}$ is the area enclosed by the $0$-level set contour $C$, ${\cal L}(\phi)=\int \mid \triangledown H(\phi({\bf x}))\mid d{\bf x}$ is the length of the $0$-level set contour $C$, $\mu$ and $\nu$ are positive weighting coefficients, and $c_1$ and $c_2$ are two constants that are used to approximate average intensities of the given image $I$ on either side of the $0$-level set contour $C$. It is obvious that $c_1$ and $c_2$ are related to the global properties of the image intensities in $\Omega_1$ and $\Omega_2$, respectively. 
This model has also been further extended to segment images into multiple parts using multiphase level set functions \citep{vese2002multiphase}. But the CV model and its multiple phase extension are both on account of the assumption that intensities of the image are statistically homogeneous in each part and use different constants to estimate intensities of these parts. They are therefore well-known as piecewise constant (PC) models, which will fail to segment images with intensity inhomogeneity when disordered intensity distribution introduces overlaps between interested objects. 

{\bf{Remark 2.}} That is to say that even though the CV model is robust to some extent with respect to noise and is also less sensitive to the initialization, it generally fails to segment images with intensity inhomogeneity \citep{chan2001active}.

\subsection{The piecewise smooth model}
In addition to introducing a local energy term as proposed in \citep{wang2010efficient} or improving original global energy by means of image local characteristics in \citep{liu2012local}, two similar ACMs were proposed by Vese and Chan \citep{vese2002multiphase} and Tsai {\emph{et al.} \citep{tsai2001curve} instead under the frame work of minimization of the Mumford-Shah functional to overcome the difficulty of the CV model in segmentation of images with intensity inhomogeneity. These models are widely known as piecewise smooth (PS) models where the image intensities are considered as two piecewise smooth functions instead of constants to represent intensities on either side of the contour $C$ \citep{vese2002multiphase} by minimizing
\begin{eqnarray}
\label{eq_PS_dataterm}
E_{PS}(u_1,u_2,\phi) &=& \sum_{i=1}^2\int\mid I({\bf x})-u_i({\bf x})\mid^2 M_i(\phi({\bf x}))d{\bf x} \nonumber \\
&+&\mu\sum_{i=1}^2\int \mid\triangledown u_i\mid ^2 M_i(\phi({\bf x}))d{\bf x} +\nu {\cal L}(\phi)
\end{eqnarray}
where $\mu$ and $\nu$ are positive weighting coefficients. 

{\bf{Remark 3.}} Although intensity inhomogeneity can be handled to some extent in the piecewise smooth model, it is obvious that the involved update of $u_1$ and $u_2$ at each iteration will certainly increase the computational burden due to solving of two partial differential equations on the entire image domain $\Omega$ \citep{liu2012local}. In addition, the level set function of the above model has to be periodically re-initialized to a signed distance function, which not only introduces problems like when and how it should be performed, but also affects numerical accuracy in an undesirable way \citep{wang2009active}.

\subsection{Region-scalable fitting model}
\label{sec_RSF}

To resolve undesirable effects caused by re-initialization, Li \emph{et al.} first introduced the following distance regularization term to intrinsically maintain the regularity of the level set function during its evolution in \citep{li2010distance} and then applied it to the region-scalable fitting (RSF) model to preserve the stability of the level set function \citep{li2008minimization}:

\begin{equation}
\label{eq_reg}
{{\cal P}}(\phi) = \int\frac{1}{2}(\mid\triangledown\phi({\bf x})\mid-1) ^2 d{\bf x}.
\end{equation} 

In \citep{li2008minimization}, local region information are incorporated into region-based level set methods relying on the assumption that intensities are locally homogeneous. Specifically, for a given point ${\bf y}\in\Omega$, two fitting functions $f_1({\bf y})$ and $f_2({\bf y})$ are used to approximate image intensities in $\Omega_1$ and $\Omega_2$, respectively. Let $\int K({\bf x}-{\bf y})\mid I({\bf x})-f_i({\bf y})\mid^2 d{\bf y}$ where $K$ is a normalized even function with the property $K({\bf u})\geq K({\bf v})$, if $\mid{\bf u}\mid<\mid{\bf v}\mid$, and ${{\rm lim}}_{\mid{\bf u}\mid\rightarrow\infty}K({\bf u})=0$. And taking all the center points ${{\bf y}}$ in the image domain $\Omega$ into consideration, the following energy functional is defined in the RSF model:
\begin{equation}
{E_{RSF}} = \int\sum_{i=1}^2 e_f^i({\bf x}) M_i(\phi({\bf x}))d{\bf x} + \nu{\cal L}(\phi) + \mu{{\cal P}}(\phi)
\end{equation}
where $\nu$ and $\mu$ are positive weighting coefficients and $e_f^i({\bf x}) = \int K({\bf x}-{\bf y})\mid I({\bf x})-f_i({\bf y})\mid^2 d{\bf y}$. 

Wang \emph{et al.} further extended the RSF model to distinguish regions with similar intensity means but different variances by introducing Gaussian distributions to describe the local image intensities \citep{wang2009active}. This improvement is in fact based on the assumption that intensities of the image obey normal distribution. Nevertheless, the image intensities are not necessarily described by a specific distribution, i.e. the intensities vary in any positions and directions and so do the intensity inhomogeneities. Therefore, histogram of the intensities and local statistics regarding the intensity and the magnitude of gradient are used to drive the evolution of the zero level set contour \citep{ni2009local,ge2012improved}.

{\bf{Remark 4.}} Although above mentioned RSF model and its improvements have shown powerful capability for segmenting images with intensity inhomogeneity, they are sensitive to the size of local scalable-region which is controlled by the kernel function $K$ and the location of the initial contour \citep{wang2015novel}. In fact, if the size of the local scalable-region is not large enough to ensure pixels inside belong to two interested objects or the zero level set contour is initialized far from the boundaries, the image will be miss-segmented or over-segmented. In addition, the above mentioned models have no capability to estimate the bias field and remove it from the inhomogenous image to be segmented.

\subsection{Local intensity clustering model}
\label{sec_LIC}
To segment images with intensity inhomogeneity and simultaneously estimate the bias field,the local intensity clustering (LIC) model was proposed based on the assumptions that 1) the bias field $b$ and the true image $J$ are multiplicative components of a given image $I$ and 2) the bias filed is slowly and smoothly varying and the true image approximately takes distinct constant values $c_1$ and $c_2$ in disjoint regions $\Omega_1$ and $\Omega_2$ \citep{li2011level}. That is to say, in a small enough circular neighborhood of a given point ${\bf y}\in\Omega$, the bias field can be seen as a constant $b({\bf y})$ and the standard K-means clustering can be used to classify intensities in the neighborhood. Taking all the center points of the entire image into account, the energy functional of the LIC model is defined by
\begin{eqnarray}
{E_{LIC}} = \int\sum_{i=1}^2 e_b^i({\bf x}) M_i(\phi({\bf x}))d{\bf x} + \nu{{\cal L}}(\phi) + \mu{{\cal P}}(\phi)
\end{eqnarray}
where $e_b^i({\bf x})=\int K({\bf x}-{\bf y})\mid I({\bf x})-b({\bf y})c_i\mid^2 d{\bf y}$, $\nu$ and $\mu$ are positive weighting coefficients, $M_1(\phi({\bf x}))$ and $M_2(\phi({\bf x}))$ are the membership functions of $\Omega_1$ and $\Omega_2$, and $K$ is a normalized function with properties described in section \ref{sec_RSF}. 

{\bf{Remark 5.}} The LIC model has shown its powerful capability for segmenting the image and estimating the bias field simultaneously \citep{li2011level}. However, drawbacks associated with the RSF model in sensitivities to the size of local scalable-region and the location of initial contour still exist in the model \citep{wang2015novel}. In addition, there is no specific constraint on the bias field to ensure its slowly and smoothly varying property. 

\subsection{Multiplicative intrinsic component optimization model}
\label{sec_MICO}
{{To constrain smoothness of the bias field explicitly, a multiplicative intrinsic component optimization (MICO) model was proposed by Li et al. in \citep{li2014multiplicative} by representing the bias field as a linear combination of $M$ basis functions $g_1$, $g_2$, ..., and $g_M$ with weighting coefficients $w_1$, $w_2$, ..., and $w_M$. In MICO, estimations of two multiplicative components, the true image that characterizes a physical property of the tissues and the bias field that accounts for inhomogeneities, are achieved simultaneously by minimizing the following fuzzy clustering energy with an iterative optimization procedure
\begin{eqnarray}
{E_{MICO}} = \int\sum_{i=1}^N \left(I({\bf x})-{\bf w}^T G({\bf x})c_i\right)^2 u_i^q({\bf x})d({\bf x})
\end{eqnarray}
where $N$ is clustering number count, $q$ is a fuzzifier to control how much clusters may overlap, $(\cdot)^T$ is the transpose operator, $G({\bf x})$ and ${\bf w}$ are column vectors defined by $G({\bf x}) = (g_1({\bf x}), g_2({\bf x}), ..., g_M({\bf x}))^T$ and ${\bf w} = (w_1,w_2, ...,w_M)^T$, respectively.}}

{{\bf{Remark 6.}} {It is obvious that MICO is in fact can be seen as an extension of the well know fuzzy c-means algorithm in supplement of bias estimation. Similar to fuzzy c-means, MICO is global clustering based and therefore sensitive to noise because of not taking into account the spatial information.}}

\subsection{Local inhomogeneous intensity clustering model}
\label{sec_LINC}
To introduce a specific constraint on the bias field and therefore ensure its estimation achieved by level set models is smoothly varying, the idea of fitting intensity biases with orthogonal basis functions is incorporated into a level set model namely Local Inhomogeneous iNtensity Clustering (LINC) in \citep{feng2016image_LINC} by defining
\begin{eqnarray}
\label{eq_LINC}
{E_{LINC}} = \int\sum_{i=1}^2 e_{{\bf w}}^i({\bf x}) M_i(\phi({\bf x})) d{\bf x} + \nu{{\cal L}}(\phi) + \mu{{\cal P}}(\phi)
\end{eqnarray}
where $e_{{\bf w}}^i({\bf x})=\int K({\bf x}-{\bf y})\mid I({\bf x})-{\bf w}^TG({\bf y})c_i\mid^2 d{\bf y}$ and all the other symbols represent the same meaning with those in section \ref{sec_LIC}.

{\bf{Remark 7.}} Due to additional two regularization terms defined on the level set function and its zero level contour, LINC is more robust to noise than MICO \citep{feng2016image_LINC}. In addition, as demonstrated in \citep{feng2016image_LINC}, the LINC model has the capability in extracting desired objects accurately from noisy images and correcting the intensity biases simultaneously, and it is robust to initialization. Furthermore, LINC converges in less iterations than RSF and LIC \citep{feng2016image_LINC}. However, convolution operation in the evolution results in a heavy computational burden.

\section{Problem formulation}
\label{sec_pf}
As well known in the literature, given an intensity inhomogeneous image $I$ defined on $\Omega$, its intensities can be viewed as
\begin{equation}
\label{eq_image_model}
I({\bf x})=b({\bf x})J({\bf x})+n({\bf x})
\end{equation}
where $I({\bf x})$ and $J({\bf x})$ are respectively the observed and true intensities at location ${\bf x}$ of the image, $b$ is the bias field accounting for the intensity inhomogeneity in the observed image, and $n$ is additive zero-mean noise \citep{vovk2007review}. In fact, the true image $J$ can be assumed to be piecewise constant that characterizes an intrinsic physical property of objects being imaged, i.e., intensity $c_i$ for the $i$-th type of objects. That is to say, the true image $J$ approximately takes $N$ distinct constant values $c_1$, $c_2$, ..., and $c_N$ in disjoint regions $\Omega_1$, $\Omega_2$, ..., and $\Omega_N$, respectively. The problem of image segmentation and bias correction is therefore considered as finding the specific intensity $c_i$ for the $i$-th type of objects and estimating the bias field $b$ at the same time.

\subsection{Representation of the bias field}
As mentioned earlier, the bias field $b$ is generally assumed to be slowly and smoothly varying in the literature. And a smooth function can be theoretically approximated by a linear combination of a given number of primary functions up to arbitrary accuracy, only if the number of the basis functions is sufficiently large \citep{powell1981approximation}. {{Therefore, as mentioned in section \ref{sec_MICO}, Li et al. represented the bias field by a linear combination of a given set of smooth primary functions $g_1$, $g_2$, ..., and $g_M$ with weighting coefficients $w_1$, $w_2$, ..., and $w_m$ in \citep{li2014multiplicative}. We follow this representation in this paper, i.e.,}}
\begin{equation}
\label{eq_basis_combination}
b({\bf x}) = \sum_{k=1}^M w_k g_k({\bf x}) = {\bf w}^T G({\bf x}).
\end{equation}
Note that the primary functions used in this paper are orthogonal and estimation of the bias field is performed by finding the optimal coefficients $w_1,w_2, ...,w_M$.

\subsection{Formulation for inhomogeneous intensity clustering}
As mentioned earlier, the true image $J$ approximately takes $N$ distinct constant values in disjoint regions $\Omega_1$, $\Omega_2$, ..., and $\Omega_N$. Therefore, taking the constant intensity $c_i$ of the true image $J$ in $\Omega_i$ into account where $i=1,2,...,N$, intensities $b({\bf x})J({\bf x})$ in this region are close to $b({\bf x})c_i$, i.e.,
\begin{equation}
b({\bf x})J({\bf x}) = b({\bf x})c_i \qquad\qquad \mbox{for} \qquad {\bf x}\in\Omega_i.
\end{equation}
Taking Eq.~(\ref{eq_basis_combination}) into account, the above equation can be rewritten as 
\begin{equation}
b({\bf x})J({\bf x}) = {\bf w}^T G({\bf x})c_i \qquad\qquad \mbox{for} \qquad {\bf x}\in\Omega_i.
\end{equation}
In consideration of the image model given in Eq.~(\ref{eq_image_model}), we have
\begin{equation}
I({\bf x}) = {\bf w}^T G({\bf x})c_i + n({\bf x}) \qquad\qquad \mbox{for} \qquad {\bf x}\in\Omega_i.
\end{equation}
As mentioned earlier, $n({\bf x})$ is additive zero-mean noise. That is to say that $\int n({\bf x})d{\bf x}=0$. Therefore, we define the following inhomogeneous intensity clustering energy
\begin{equation}
\label{eq_iice}
{\cal F}=\sum_{i=1}^{N}\lambda_i\int_{\Omega_i}(I({\bf x})-{\bf w}^T G({\bf x})c_i)^2 d{\bf x}
\end{equation}
where $\lambda_1, \lambda_2, ..., \lambda_N$ are positive constants to indicate preference of the proposed model to corresponding classes. Note that when boundaries of the regions $\Omega_i$ for $i=1,2,...,N$ are consistent with reality, i.e., they locate exactly at right boundaries of the desired objects, the above defined energy takes its minimal value. 

{\bf{Remark 8.}} {{Note that the energy defined above is distinct from MICO proposed in \citep{li2014multiplicative}. First, we estimate the true image directly by piecewise constant functions $J({\bf x})=c_i$ for ${\bf x}\in\Omega_i$ where $i=1,2,...,N$, whereas fuzzy membership functions with a predefined real exponent $q$ are included in MICO to represent regions $\Omega_i$. Second, we combine the above defined energy with a region based level set model as given in the next subsection which generates hard segmentation with memberships of $\Omega_i$ represented by level set functions, whereas MICO is intensity globally clustering based with fuzzy memberships computed immediately from image intensities and clustering centroids.}}

\subsection{Two phase level set formulation $IEOPF^2$}
It is obvious that the proposed energy in Eq.~(\ref{eq_iice}) is expressed in terms of the regions $\Omega_1$, $\Omega_2$, ..., and $\Omega_N$, which makes it difficult to derive a solution to minimize the energy from this expression. In the case that the image domain $\Omega$ is separated into two disjoint regions $\Omega_1$ and $\Omega_2$, i.e., $N=2$, the energy defined in Eq.~(\ref{eq_iice}) can be converted to a level set formulation by representing the two disjoint regions with a given level set function $\phi$ defined on $\Omega$. Then, the energy minimization problem can be solved by using well-established variational methods \citep{li2011level}. Let the level set function $\phi$ take negative and positive signs on either side of the 0-level set contour denoted by $C \triangleq \lbrace {\bf x}: \phi({\bf x})=0\rbrace$, which can be used to represent a partition of the domain $\Omega$ with two disjoint regions. The disjoint regions separated by the contour can be represented by $\Omega_1 \triangleq \lbrace {\bf x}: \phi({\bf x})<0\rbrace$ and $\Omega_2 \triangleq \lbrace {\bf x}: \phi({\bf x})>0\rbrace$. In consideration of properties of the Heaviside function $H$, the regions are further represented by the following member functions $M_{1}(\phi({\bf x}))=1-H(\phi({\bf x}))$ and $M_{2}(\phi({\bf x}))=H(\phi({\bf x}))$, respectively. Thus, for the case $N=2$, we rewrite the energy ${\cal F}$ described in Eq.~(\ref{eq_iice}) into the following level set formulation
\begin{equation}
\label{eq_img_term}
{\cal F}=\sum_{i=1}^{2}\lambda_i\int(I({\bf x})-{\bf w}^T G({\bf x})c_i)^2 M_{i}(\phi({\bf x}))d{\bf x}
\end{equation}
It is obvious that the energy ${\cal F}$ is a functional of variables the level set function $\phi$, the vector ${\bf c} =(c_1,c_2)^T$, and the weight coefficients of the basis functions ${\bf w} = (w_1,w_2, ...,w_M)^T$, i.e., ${\cal F}(\phi, {\bf c}, {\bf w})$. The energy ${\cal F}(\phi, {\bf c}, {\bf w})$ is the data
term of the final energy functional of the proposed level set formulation, defined by
\begin{equation}
\label{eq_energy}
E_{}(\phi, {\bf c}, {\bf w})={\cal F}(\phi, {\bf c}, {\bf w}) + \nu{{\cal L}}(\phi) + \mu{{\cal P}}(\phi)
\end{equation}
where ${\cal P}$ is the regularization term defined in Eq.~(\ref{eq_reg}) used here to maintain the regularity of the level set function $\phi$ and ${\cal L}$ is the same arc length term used in state-of-the-art models to smooth the $0$-level set contour.

{\bf Remark 9.} {{The proposed model is essentially different from MICO which is in fact a global clustering method that can be seen as an extension of fuzzy c-means in bias correction \citep{li2014multiplicative} and is therefore sensitive to noise \citep{feng2016image_LINC}. But the proposed model is level set based by introducing the idea of basis function fitting proposed in MICO into the energy formulation and the regularization terms in the proposed model can suppress to some extent negative effects of noise. In addition, the above defined model separates images into two parts and we will extend it into multichannel and multiphase patterns to segment colorful images with multiple objects in the next two subsections, whereas MICO is only suitable for segmenting gray images into parts, the number of which is predefined.}} On the other hand, the proposed model is also different from LIC and LINC. First, there is no normalized even convolution kernel function in the proposed model and the integral is therefore one layer which is less than either LIC or LINC. Second, in the proposed method, an explicit constraint on estimation of the bias field is introduced to ensure the slowly and smoothly varying property of the bias field compared with LIC. Relationship of the proposed model with CV and PS will be discussed in Section \ref{sec_relatewother}.

\subsection{Extension to multichannel case $IEOPF^2_L$}
It is obvious that the above model defined in Eq.~(\ref{eq_energy}) is applicable in extracting interested objects from gray images. But multichannel images of the same scene that come from different imaging modalities or color images are becoming more and more common in our life. To extend the proposed model to be able to extract interested objects from multichannel images, we first denote a given multichannel image ${\bf I}$ by ${\bf I}=(I_1,I_2,...,I_L)$ where $L$ is the channel number of ${\bf I}$. Let $e_i({\bf x})=\sum_{j=1}^{L} \gamma_j\left( I_j({\bf x})-{{\bf w}_j}^T G({\bf x})c_{ij}\right) ^2$ where $\gamma_j$ are positive weighting coefficients that are used to control influence of the $j$-th channel. We then rewrite Eq.~(\ref{eq_img_term}) as follows
\begin{equation}
\label{eq_colorimg_term}
{\cal F}(\phi, {\bf C}, {\bf W})=\sum_{i=1}^{2}\lambda_i\int e_i({\bf x}) M_{i}(\phi({\bf x}))d{\bf x}
\end{equation}
where ${\bf C}$ is an $2 \times L$ matrix defined by ${\bf C}=({\bf c}_1,{\bf c}_2,...,{\bf c}_L)$ and ${\bf W}$ is a matrix with $M \times L$ elements defined by ${\bf W}=({\bf w}_1,{\bf w}_2,...,{\bf w}_L)$. We finally rewrite Eq.~(\ref{eq_energy}) as follows
\begin{equation}
\label{eq_multichannel_energy}
E_{}(\phi, {\bf C}, {\bf W})={\cal F}(\phi, {\bf C}, {\bf W}) + \nu{{\cal L}}(\phi) + \mu{{\cal P}}(\phi).
\end{equation}

\subsection{Further extension to multiphase case $IEOPF^N_L$}
Since one level set function $\phi$ can only be used to represent 2 subregions of image domain $\Omega$ denoted by membership functions $M_1$ and $M_2$, which are in fact inside and outside of the zero level contour of $\phi$, $Q$ level set functions are required to represent $N$ subregions where $Q=\lceil log_2(N)\rceil$. Thus, the subregion $\Omega_i$ can be represented by the member function $M_i(\Phi)$, i.e., $M_i(\Phi({\bf x}))=1$ for ${\bf x}\in \Omega_i$ and $M_i(\Phi_1({\bf x}))=0$ otherwise where $\Phi=(\phi_1,\phi_2,...,\phi_K)$ and $i=1,2,...,N$. To extend the proposed model to segment multiple objects from images with intensity inhomogeneity, we first further rewrite Eq.~(\ref{eq_colorimg_term}) as follows
\begin{equation}
\label{eq_finalimg_term}
{\cal F}(\Phi, {\bf C}, {\bf W})=\sum_{i=1}^{N}\lambda_i\int e_i({\bf x}) M_{i}(\Phi({\bf x}))d{\bf x}.
\end{equation}
We then define ${\cal  P}(\Phi)=\sum_{q=1}^{Q}{\cal P}(\phi_q)$ and ${\cal  L}(\Phi)=\sum_{q=1}^{Q}{\cal L}(\phi_q)$ where ${{\cal P}}(\phi_q) = ({1}/{2})\int(\mid\triangledown\phi_q({\bf x})\mid-1) ^2 d{\bf x}$ and ${\cal L}(\phi_q)=\int \mid \triangledown H(\phi_q({\bf x}))\mid d{\bf x}$, respectively. Finally, we rewrite Eq.~(\ref{eq_multichannel_energy}) as follows
\begin{equation}
\label{eq_multiphase_energy}
E_{}(\Phi, {\bf C}, {\bf W})={\cal F}(\Phi, {\bf C}, {\bf W}) + \nu{\cal  L}(\Phi) + \mu{\cal  P}(\Phi).
\end{equation}

\subsection{Energy minimization}
In the proposed model, segmentation and bias correction are determined by the final level set function $\hat{\Phi}$ and the optimal weighting coefficients $\hat{\bf W}$ that are obtained by minimizing the energy functional $E_{}(\Phi, {\bf c}, {\bf w})$ defined in Eq.~(\ref{eq_multiphase_energy}). The energy minimization is achieved by an iterative process. That is to say, the energy functional $E_{}(\Phi, {\bf C}, {\bf W})$ is minimized with respect to each of its variables $\Phi$, ${\bf C}$, and ${\bf W}$ in each iteration by fixing the other two with values from last iteration. 

For fixed ${\bf C}$ and ${\bf W}$, we minimize the energy functional $E_{}(\Phi, {\bf C}, {\bf W})$ with respect to $\Phi=(\phi_1,\phi_2,...,\phi_Q)$ using the standard gradient descent method and obtain
\begin{eqnarray}
\label{eq_iter_phi}
\frac{\partial\phi_q}{\partial t}=-\sum_{i=1}^{N}\frac{\partial M_i(\Phi)}{\partial\phi_q}\lambda_i e_i +\mu \left(\triangledown^2\phi_q - {{\rm div}}\left(\frac{\triangledown\phi_q}{\mid\triangledown\phi_q\mid}\right)\right)\nonumber \\
+\nu \delta(\phi_q){{\rm div}}\left(\frac{\triangledown\phi_q}{\mid\triangledown\phi_q\mid}\right)
\end{eqnarray}
where $q=1,2,...,K$.

For fixed $\Phi$ and ${\bf W}$, we minimize the energy functional $E_{}(\Phi, {\bf C}, {\bf W})$ with respect to ${\bf C}$ by solving the equation $\frac{\partial E}{\partial {\bf C}}={\bf 0}$ where ${\bf 0}$ is a $N \times L$ matrix with constant value $0$ and obtain
\begin{equation}
\label{eq_iter_c}
c_{ij} = \frac{\int \left(I_{j}({\bf x}){{\bf w}_j}^T G({\bf x})\right)M_i(\Phi({\bf x})) d{\bf x}}{\int \left({{\bf w}_j}^T G({\bf x})\right)^2 M_i(\Phi({\bf x}))d{\bf x}}
\end{equation}
where $i=1,2,...,N$ and $j=1,2,...,L$.

For fixed $\Phi$ and ${\bf C}$, we minimize the energy functional $E_{}(\Phi, {\bf C}, {\bf W})$ with respect to  ${\bf W}$ by solving the equation $\frac{\partial E}{\partial {\bf W}}={\bf 0}$ where ${\bf 0}$ is a $M \times L$ matrix with constant value $0$ and obtain
\begin{equation}
\label{eq_iter_w}
{\bf w}_j=A_j^{-1}{\bf v}_j
\end{equation}
where $j=1,2,...,L$ and $A_j$ is a matrix with $M \times M$ elements and ${\bf v}$ is an $M$-dimensional column vector, given by
\begin{equation}
\label{eq_minimize_wA}
A_j = \int{\left(\sum_{i=1}^N{\lambda_ic_{ij}^2M_i(\phi({\bf x}))}\right)G({\bf x})G^T({\bf x})}d{\bf x}
\end{equation}
and 
\begin{equation}
\label{eq_minimize_wv}
{\bf v}_j = \int{\left(I_j({\bf x})\sum_{i=1}^N{\lambda_ic_{ij}M_i(\phi({\bf x}))}\right)G({\bf x})}d{\bf x}.
\end{equation}

{\bf Remark 10.} {{Although minimizations of the proposed model with respect to its variables are similar to MICO, the differences are given as follows. Memberships of the proposed model are implicited in level set functions. Therefore, we first minimize of the energy formulation with respect to level set functions using gradient descent method and then compute hard memberships with updated level sets instead of computing fuzzy memberships directly from the energy formula defined in MICO with partial differential method. As we have extended the proposed model to segment multichannel images, we estimate a bias field and $N$ distinct constants for each image channel and therefore minimize the energy with respect to each of the variables instead totally estimating one bias and $N$ distinct cluster centroids in MICO.}}

\subsection{Implementation}
\label{sec_imp}
In our numerical implementation, the Heaviside function $H$ is approximated by a smooth version $H_\epsilon$ with $\epsilon = 1$, most popularly used in the literature \citep{chan2001active,vese2002multiphase,li2008minimization,li2010distance,li2011level,feng2016image_LINC}, defined by
\begin{equation}
\label{eq_heaviside_approx}
H_\epsilon(x) = \frac{1}{2}\left[1+\frac{2}{\pi}\arctan\left(\frac{x}{\epsilon}\right)\right].
\end{equation}
The derivative of $H_\epsilon$ is used to approximate the Dirac delta function $\delta$, which can be written as
\begin{equation}
\label{eq_delta_approx}
\delta_\epsilon(x) = H^\prime_\epsilon(x) = \frac{1}{\pi}\frac{\epsilon}{\epsilon^2+x^2}.
\end{equation}

In this paper,  10 orthogonal Legendre polynomial functions, which are four order precision, are used to approximately estimate the bias field, i.e., $M=10$. In fact, for each ${\bf x}\in\Omega$, we can rewrite ${\bf x}$ as ${\bf x}=(x_1,x_2)$ where $x_1$ and $x_2$ are directional components of the given two-dimensional image $I$ defined on $\Omega$. The smooth basis functions $g_1$, $g_2$, ..., and $g_{15}$ used in this paper are defined by $g_1({\bf x})=1$, $g_2({\bf x})=x_1$, $g_3({\bf x})=(3x_1^2-1)/2$, $g_4({\bf x})=(5x_1^3-3x_1)/2$, $g_5=x_2$, $g_6=x_1 x_2$, $g_7({\bf x})=(3x_1^2-1)x_2/2$, $g_{8}({\bf x})=(3x_2^2-1)/2$, $g_{9}({\bf x})=x_1(3x_2^2-1)/2$, $g_{10}({\bf x})=(5x_2^3-3x_2)/2$. The column vector $G({\bf x})$ can be therefore written as $G({\bf x})=(g_1({\bf x}),g_2({\bf x}),...,g_{10}({\bf x}))^T$. The implementation of the proposed model can be straightforwardly expressed as follows in {\bf Algorithm 1}.

\begin{algorithm}
\caption{The proposed bias correction embedded level set model IEOPF}
\label{alg_BCELS}
\begin{algorithmic}[1]
\Require
The multichannel image ${\bf I}$, its channel number $L$, and the number of interested objects $N$.
\Ensure
Segmentation results determined by membership function $M_i(\Phi)$ and the bias field ${\bf b}=(b_1,b_2,...,b_L)$ with each $b_j$ estimated by ${{\bf w}_j}^T G({\bf x})$ where $i=1,2,...,N$ and $j=1,2,...,L$.

\State
Initialize ${\bf W}$ with a random $M \times L$ matrix and $\phi_k$ with a binary step function, defined by $\phi_q({\bf x})=-a$ for ${\bf x}$ inside the initial zero-level contour of $\phi_q$ and $\phi_q({\bf x})=a$ otherwise, where $q=1,2,...,Q$.

\State Update cluster center matrix ${\bf C}$ with its elements $c_{ij}$ computed using Eq.(\ref{eq_iter_c}) where $i=1,2,...,N$ and $j=1,2,...,L$.

\State
Update $\phi_q$ by adding it with the difference determined by post-multiplying Eq.(\ref{eq_iter_phi}) with $\Delta t$ where $\Delta t$ represents the step of temporal difference and $q=1,2,...,Q$.

\State
Update the weighing coefficients matrix ${\bf W}$ with each column of ${\bf W}$ computed using Eq.(\ref{eq_iter_w}).

\State
Check convergence criterion and iteration number. If convergence has been reached or the iteration number exceeds a predetermined maximum number, stop the iteration, otherwise, go to Step 2.

\end{algorithmic}
\end{algorithm}

Note that the convergence criterion used in this paper is $\sum_{i=1}^{N}\sum_{j=1}^{L}\parallel {{c_{ij}}}^{(n+1)} - {{c_{ij}}}^{(n)} \parallel _2 < 0.001$, where ${{c_{ij}}}^{(n)}$ is the cluster center ${{c_{ij}}}$ updated at the $n$-th iteration and $\parallel\star\parallel _2$ is the Euclidean distance of $\star$.

{\bf{Remark 11.}} The main additional computational cost in the proposed model is for computing ${\bf w}_j$ in Eq.~(\ref{eq_iter_w}) compared with state-of-the-art models reviewed in Section \ref{sec_relatedworks}. However, we notice that $G({\bf x})$ and $I_j({\bf x})$ are independent of the level set functions $\Phi$ and clustering centers ${\bf C}$ which indicate that we can compute $G({\bf x})G^T({\bf x})$ for Eq.~(\ref{eq_minimize_wA}) and $I_j({\bf x})G({\bf x})$ for Eq.~(\ref{eq_minimize_wv}) in advance and keep the results fixed during the iteration to accelerate the proposed model.

\section{Experimental results}
\label{sec_expresult}

We have tested the proposed model extensively on synthetic and real images in Matlab R2016a on a computer with Intel(R) Core(TM)i5-3230M 2.6GHzCPU,4GBRAM,and Windows7 64-bit operating system. In this section, we first evaluate effectiveness of the proposed model IEOPF on synthetic images that are widely used to verify ACMs and selected natural images from public datasets. We then evaluate the proposed model on two pubic MR brain image repositories qualitatively and quantitatively. Unless otherwise specified, we set $a=2.0$, $\Delta t=0.1$, $\lambda_1=\lambda_2=\lambda_3=1.0$, $\mu=1.0$, and $\nu=0.005\times 255 \times 255$ in this paper.

\subsection{Effectiveness of IEOPF}
In this subsection, we qualitatively evaluate effectiveness of the proposed model on synthetic images and selected natural images from public datasets and give the validation results in the following paragraphs. Note that the synthetic and natural images are either widely used in the literature to verify active contour models or appropriate for application of the proposed model IEOPF.

We first apply the proposed model to three synthetic gray images (\emph {widely used to evaluate active contour models in the literature}), one cardiac X-ray image, and one brain MR image in this subsection. It is obvious that segmentation results of the proposed model on the images are agreed with contents contained in the images even though intensities of the images are not homogeneous due to existing of severe intensity biases as shown in Fig.~\ref{fig:effect1}. That is to say that it is difficult to extract interested objects from the images because intensity ranges of objects (including the background) in the images are overlapped due to severe intensity inhomogeneities existed in the images which manifests as there are no well-separated peaks in intensity histograms of the images as shown in Fig.~\ref{fig:effect1}. However, there are well-defined and separated peaks in histograms of the bias corrected images, each corresponding to one object or the background. This demonstrates the capability of the proposed model in correcting bias fields from images with intensity inhomogeneity. Meanwhile, the biases estimated by the proposed model with orthogonal primary functions are all slowly (not sharply) varying as shown in Fig.~\ref{fig:effect1} which meets properties of the bias field as described in section \ref{sec:intro}.

\begin{figure}[!htb]
\centering
{\includegraphics[width=0.75\textwidth]{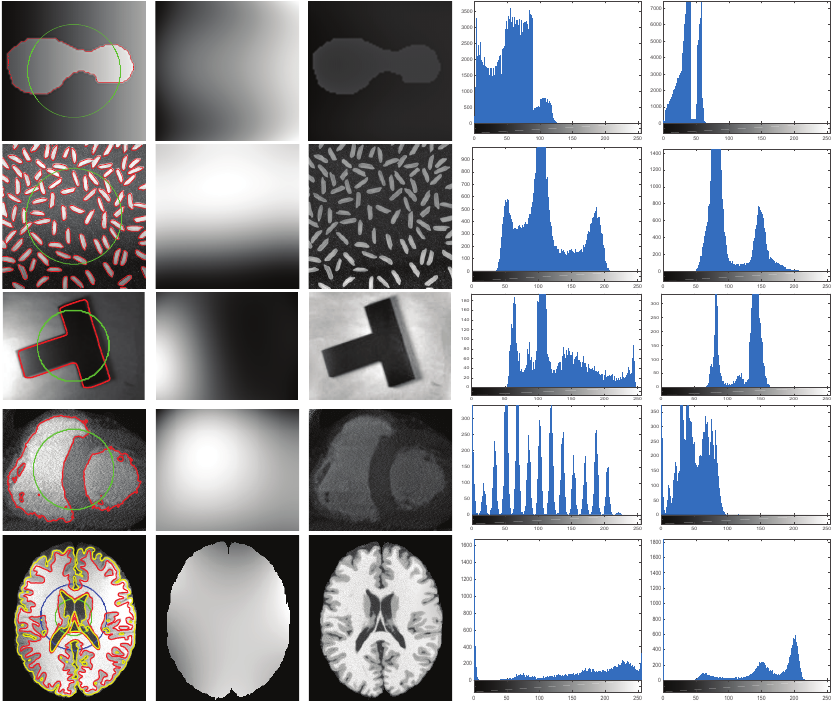}}
\\
\scriptsize\leftline{\qquad\qquad\qquad\quad Orig w/ Init \& Seg \qquad\quad Bias \qquad\qquad\quad Corrected \qquad\quad Orig Histogram \qquad Corrected Histogram}
\caption{Results of IEOPF in segmenting inhomogeneous gray images and correcting the biases. \label{fig:effect1}}
\end{figure}

We then apply the proposed two phase level set model to segment four selected natural images with three color channels from BSD database \citep{amfm_pami2011}, namely 135069, 42049, 3096, and 86016, respectively. \emph {The reason we selected these images is that each of the images contains only one object besides the background which can therefore be distinguished with one level set function}. Results of the proposed model on segmentation of the images with two phase level sets are given in Fig.~\ref{fig:effect2}. It is obvious that the estimated biases are smoothly varying and the corrected images are more homogeneous than the originals. Furthermore, the energy functional of the proposed model defined in Eq.(\ref{eq_multichannel_energy}) is converged (generally in less than 50 iterations) as shown in the right column of Fig.~\ref{fig:effect2}.

\begin{figure}[!h]
\centering
{\includegraphics[width=0.75\textwidth]{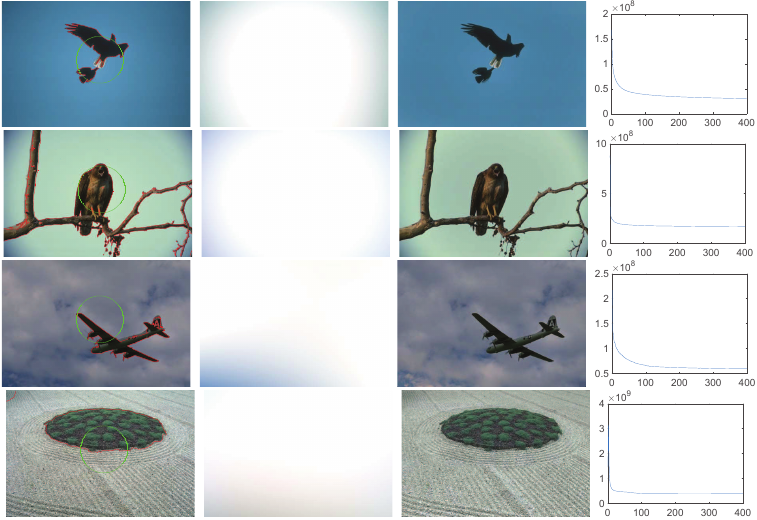}}
\\
\scriptsize\leftline{\qquad\qquad\qquad\qquad Orig w/ Init \& Seg \qquad\qquad\qquad Bias \qquad\qquad\qquad\qquad Corrected \quad\qquad\qquad Energy Curve}
\caption{Results of IEOPF$_3^2$ in segmenting natural images from BSD database.\label{fig:effect2}}
\end{figure}

We thirdly apply the proposed three phase level set model to segment two MR brain images which are corrupted by severe intensity inhomogeneities and two selected natural images from MSRCORID database \citep{MSRCORID}, namely, 164\_6484 and 112\_1204. \emph {The first two images are widely used to evaluate multiple phase active contour models in the literature and the last two images are selected because three kind of objects are contained which are suitable for three phase segmentation.} From the results given in Fig.~\ref{fig:effect3}, we can see that the estimated biases are smooth and the corrected images are much more homogeneous. Moreover, the extracted objects are coincided with the images.

\begin{figure}[!h]
\centering
{\includegraphics[width=0.75\textwidth]{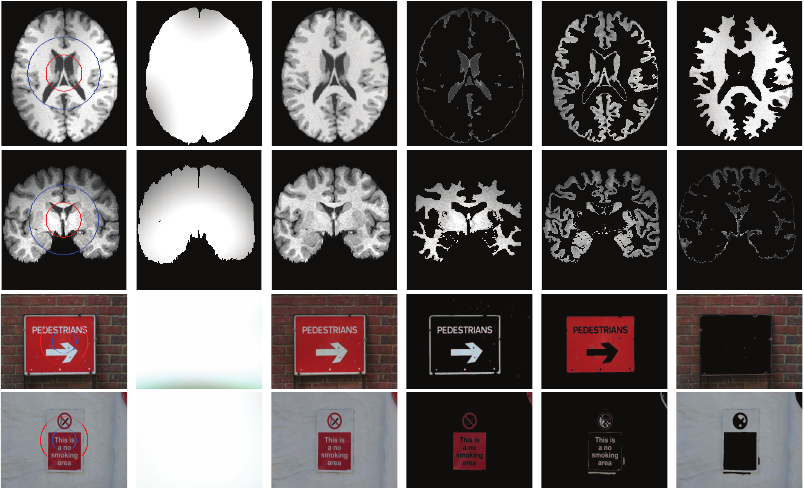}}
\\
\scriptsize\leftline{\qquad\qquad\qquad\qquad Orig w/ Init \qquad\qquad Bias \qquad\quad Corrected \qquad\qquad Seg1 \qquad\qquad\quad Seg2 \qquad\qquad\quad Seg3}
\caption{Results of IEOPF$_1^3$ and IEOPF$_3^3$ in segmeting brain MR images and natural images from MSRCORID database, respectively. \label{fig:effect3}}
\end{figure}

We fourthly evaluate energy convergence of the proposed model on all above mentioned images and show iteration process of the proposed model on four of them in Fig.~\ref{fig:effect4}. \emph {The images are appropriate to evaluate the proposed model in the sense of one-channel-two-phase, one-channel-multiple-phase, multiple-channel-two-phase, and multiple-channel-multiple-phase, respectively.} It can be seen that the proposed model is convergent and satisfactory results can be generally obtained in less than 20 iterations. Note that three kinds of color are used to show the results clearly.

\begin{figure}[!h]
\centering
{\includegraphics[width=0.75\textwidth]{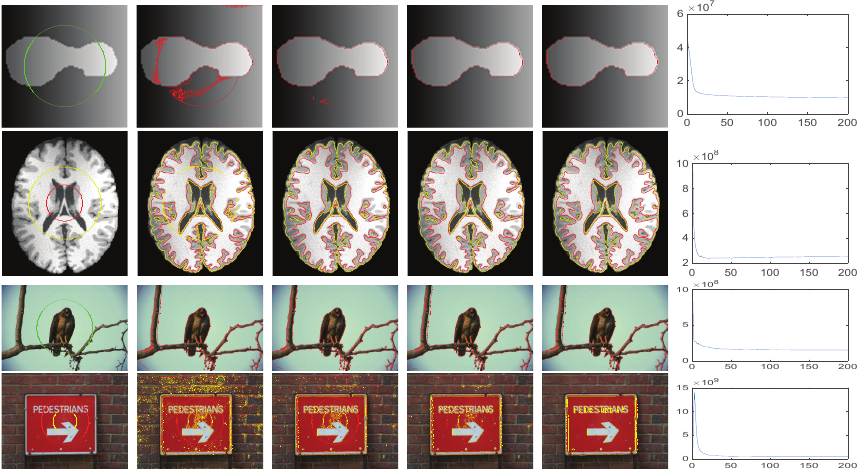}}
\\
\scriptsize\leftline{\qquad\qquad\qquad\qquad Orig w/ Init \qquad\quad 5 Iters \qquad\quad 10 Iters \qquad\quad 15 Iters \qquad\quad Final \qquad\qquad Energy Curve}
\caption{Demonstration of $0$-level contour evolutions of the proposed model IEOPF.\label{fig:effect4}}
\end{figure}

We finally compare results of the proposed model with state-of-the-art models on one synthetic image and one natural image from BSD qualitatively and show the result in Fig.~\ref{fig:cmp1}. \emph {Note that to be fair, initializations on either image are all the same for each of the comparable models.} And we set the parameters $\lambda_1=\lambda_2=1.0$, $\mu=1.0$, and $\nu=0.005\times255\times255$. The only parameter of MICO, fuzzy qualifiers, is set to be $2$. As the CV, RSF, LIC, LINC, and MICO models are short of the capability to extract interested objects from color images directly, we first convert the color image to a gray image using the rgb2gray function of matlab and then input the image to the models. However, the proposed model can be directly used to deal with color images (three channels). Therefore, segmentation contour of the proposed model on the natural image given is marked on the original colourful image whereas results of the others are marked on the gray images. As shown in Fig.~\ref{fig:cmp1}, due to absence of dealing with intensity inhomogeneity, segmentation results of the CV model include other regions besides geometrical shapes really exist in the synthetic image and eagles in the natural image. Segmentation results of the RSF model are a little better than those of the CV model because it can handle intensity inhomogeneity to some extent. But the RSF model lacks the capability of bias estimation and correction. As shown in Fig.~\ref{fig:cmp1}, the bias fields estimated by the LIC model are obviously not smooth enough and segmentation results are certainly wrong. {{However, bias estimated by MICO is smoother than LIC due to basis functions used to fit inhomogeneities. But there are over segmentations at corners of the images as shown in Fig.~\ref{fig:cmp1}.}} Although segmentation results and bias estimations of the LINC model are desirable, color images can not be directly input into the model before being converted to gray ones. In addition, as mentioned in section \ref{sec:intro}, convolution operation in the evolution results in a heavy computational burden for LINC which we will further discuss in section \ref{sec:cmp2LINC}. It obvious that the proposed model achieves the best segmentations, bias estimations and corrections.

\begin{figure}[!h]
\centering
{\includegraphics[width=0.75\textwidth]{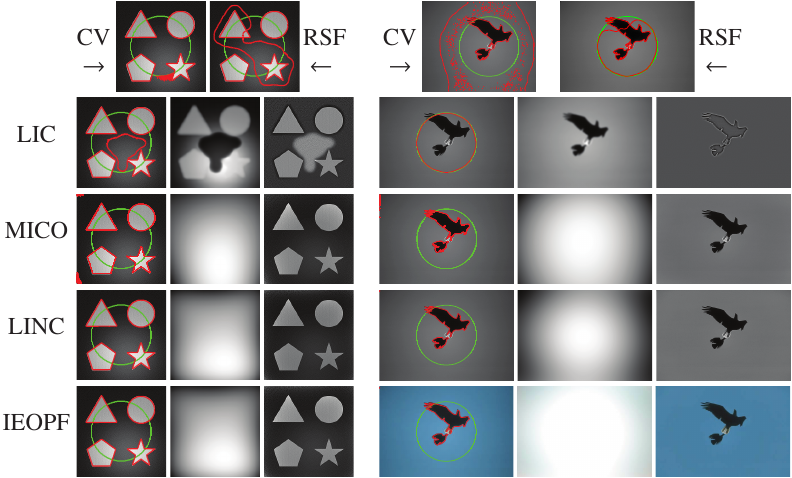}}
\\
\scriptsize\leftline{\qquad\qquad\qquad\qquad\qquad\qquad Init \& Seg \quad Bias \qquad Corrected \qquad\quad Init \& Seg \qquad\qquad Bias \qquad\qquad Corrected}
\caption{Qualitative comparison with state-of-the-art models on one synthetic image (left) and one naturnal image from BSD (right). \label{fig:cmp1}}
\end{figure}

\subsection{Evaluation on pubic image repositories}
In this subsection, we evaluate effectiveness of the proposed model quantitatively on one simulated MR dataset and one real MR image dataset. The first one consists of 9 cases of MR images with three different levels of noise and intensity inhomogeneity, respectively. Resolutions of the images are $181\times 217\times 181$ with 1 mm in-plane pixel size and 1 mm slice thickness. For more information about the dataset, interested readers are referred to the website \url{http://brainweb.bic.mni.mcgill.ca/brainweb/} and the reference \citep{cocosco1997brainweb}. To construct a much more challenging dataset for segmentation methods, three more levels of non-linear intensity inhomogeneities are added to the original image with noises. Therefore, there are totally 18 image cases for the first image dataset. The second image set is known worldwide as the Internet Brain Segmentation Repository (IBSR) which contains 18 cases of
T1-weighted brain MR image cases with skull-removed masks and manually-guided expert segmentation results. Resolutions of the images are all $256\times128\times256$. Interested readers are referred to \url{https://www.nitrc.org/projects/ibsr} for detail. Note that for each image case, the segmentation task is to extract white matter (WM), gray matter (GM), and cerebrospinal fluid (CSF) from the background. As intensities of the background are all zero for the images, two level set functions are used to partition the images into three regions that is $K=2$ and $N=3$. To compare performance of the proposed model with state-of-the-art models like CV, LIC, and LINC on these image datasets, we first extend the comparable models to three phase ({\emph{matlab codes will be released on our personal homepage if this paper got published}}). We then define membership functions $M_1=(1-H(\phi_1))(1-H(\phi_2))$, $M_2=(1-H(\phi_1))H(\phi_2)$, and $M_3=H(\phi_1)$ to represent WM, GM, and CSF, respectively. For a fair comparison, we first extend comparison models to three phase and then use the same parameter set and the same strategy to initialize the level set functions for all the comparison models. The initialization strategy is that areas separated by a predefined threshold are adopted to initialize $\phi_1$ by considering the areas as inside and outside of the zero level contour. Areas separated by another predefined threshold are adopted to initialize the level set function $\phi_2$. The thresholds are adaptively defined as 0.8 and 0.3 times of maximal intensity of pending to be segmented images. We have to point out that the proposed model is robust to initialization which will be discussed in section \ref{sec:initrobust}. Note that we applied the proposed model and comparable state-of-the-art models only on image slices that really contain WM, GM, and CSF.

\subsubsection{Qualitative comparison}
\label{sec:qualcomp}
Segmentation results of the proposed model with three state-of-the-art level set models, i.e., CV, LIC, and LINC, and the global clustering based MICO on the $90$-th slice of selected brainweb cases and the $128$-th slice of selected IBSR image cases are given in Fig.~{\ref{fig:bwimgcmp}} and Fig.~{\ref{fig:ibsrimgcmp}}. The corresponding bias estimation and correction results are given in Fig.~{\ref{fig:bwbiascmp}} and Fig.~{\ref{fig:ibsrbiascmp}}, respectively. The reason we select these image is that they are the most noisy and biased and they are therefore challengeable. It can be seen that 1) the proposed model is much more robust to noises and bias fields and 2) segmentation results of the proposed model are much more close to corresponding ground truth. {{Due to potential relatedness of the proposed model to MICO, it is necessary to compare them qualitatively and quantitatively, beside describing theoretical differences as given in Remarks 8-10. It can be obviously seen that 1) MICO is sensitive to noise, especially as shown for the first two images of Fig.~{\ref{fig:bwimgcmp}} with 9\% and 6\% noises to the brightest tissue and 2) MICO prefers to provide high biases at image centres which can be seen from Fig.~{\ref{fig:ibsrimgcmp}} and especially for the first image in Fig.~{\ref{fig:bwimgcmp}} with no intensity biases actually. But bias fields estimated by the proposed model are much more matching with the actual situation and the bias estimated is almost a constant for the first image in Fig.~{\ref{fig:bwimgcmp}} which is not corrupted by inhomogeneities in fact.}} Quantitative evaluation will be given in section \ref{sec:quaneval}.

\begin{figure}[!h]
\centering
{\includegraphics[width=0.75\textwidth]{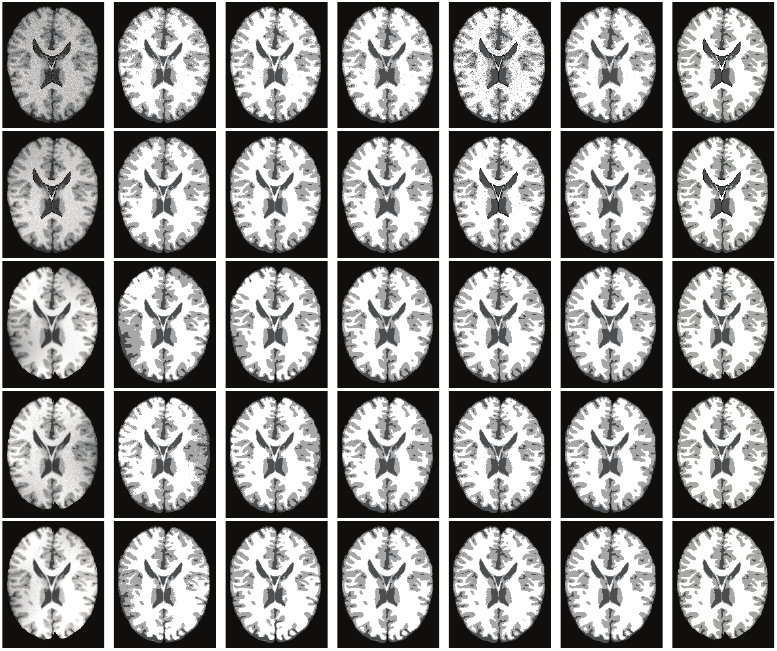}}
\\
\scriptsize\leftline{\qquad\qquad\qquad\qquad\qquad Orig \qquad\qquad CV \qquad\qquad LIC \qquad\qquad LINC \qquad\qquad MICO \qquad\quad IEOPF$_1^3$ \qquad\quad GT}
\caption{Segmentation comparison with state-of-the-art models on selected BrainWeb images.\label{fig:bwimgcmp}}
\end{figure}

\begin{figure}[!h]
\centering
\scriptsize\leftline{\qquad\qquad\qquad\qquad\quad Orig \qquad\qquad\qquad\qquad\qquad\quad Bias \qquad\qquad\qquad\qquad\qquad\qquad\qquad\qquad Corrected}
{\includegraphics[width=0.75\textwidth]{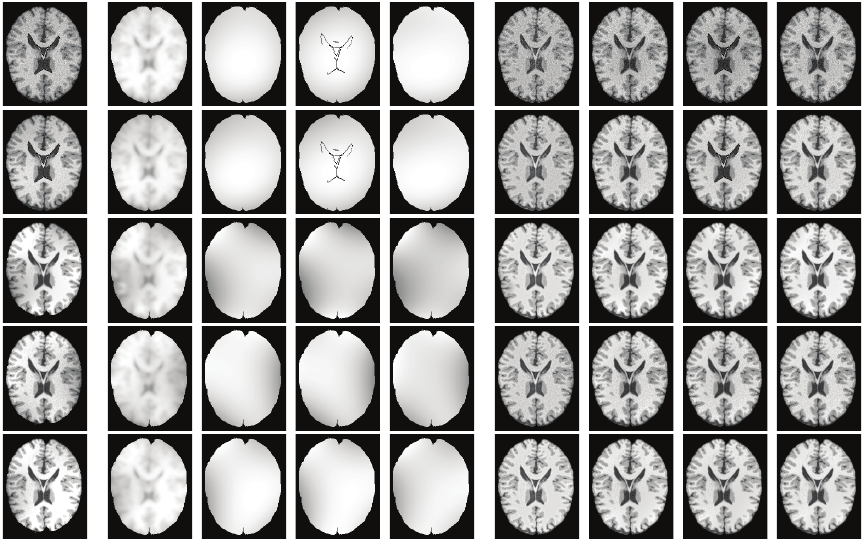}}
\\
\scriptsize\leftline{\qquad\qquad\qquad\qquad\qquad\qquad\qquad LIC \qquad LINC \qquad MICO \qquad IEOPF$_1^3$ \quad\quad LIC \qquad LINC \qquad MICO \qquad IEOPF$_1^3$ }
\caption{Comparison of bias estimation and correction with state-of-the-art methods on selected BrainWeb images.\label{fig:bwbiascmp}}
\end{figure}

\begin{figure}[!h]
\centering
{\includegraphics[width=0.75\textwidth]{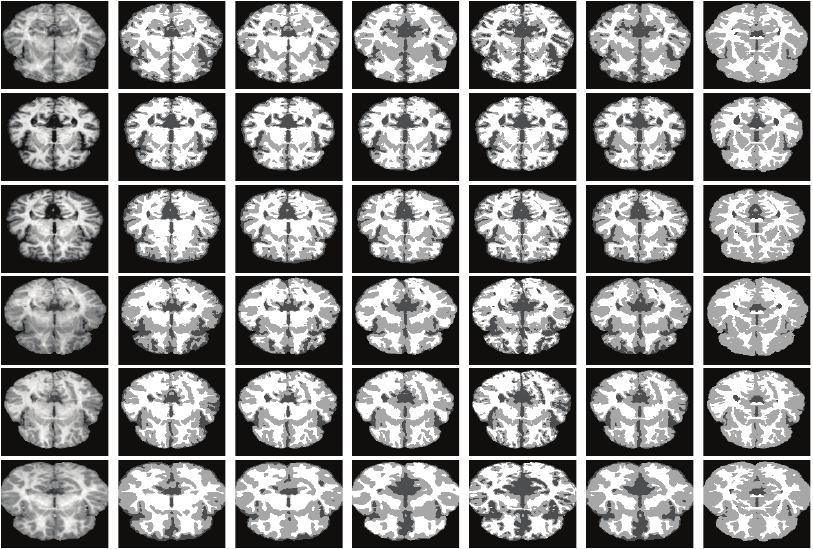}}
\\
\scriptsize\leftline{\qquad\qquad\qquad\qquad\qquad Orig \qquad\qquad CV \qquad\qquad LIC \qquad\qquad LINC \qquad\qquad MICO \qquad\quad IEOPF$_1^3$ \qquad\quad GT}
\caption{Segmentation comparison with state-of-the-art methods on selected IBSR images.\label{fig:ibsrimgcmp}}
\end{figure}

\begin{figure}[!h]
\centering
\centering
\scriptsize\leftline{\qquad\qquad\qquad\qquad\quad Orig \qquad\qquad\qquad\qquad\qquad\quad Bias \qquad\qquad\qquad\qquad\qquad\qquad\qquad\qquad Corrected}
{\includegraphics[width=0.75\textwidth]{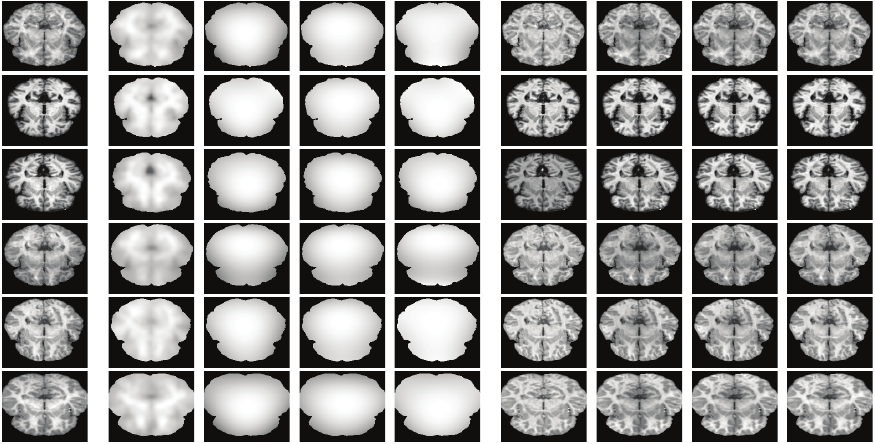}}
\\
\scriptsize\leftline{\qquad\qquad\qquad\qquad\qquad\qquad\qquad LIC \qquad LINC \qquad MICO \qquad IEOPF$_1^3$ \quad\quad LIC \qquad LINC \qquad MICO \qquad IEOPF$_1^3$ }
\caption{Comparison of bias estimation and correction with state-of-the-art methods on selected IBSR images.\label{fig:ibsrbiascmp}}
\end{figure}

\subsubsection{Quantitative evaluation}
\label{sec:quaneval}

To quantitatively evaluate segmentation results of the proposed framework with state-of-the-art methods, false positive ratio (FPR), false negative ratio (FNR), and dice similarity coefficient (DSC) are metrics used in this paper. Let NFP and NFN be the number of FP (false positive) and FN (false negative) and A be the ground truth, FPR and FNR can then be defined by
\begin{equation}
FPR = \frac{NFP}{|I|-|A|}
\end{equation}
and
\begin{equation}
FNR = \frac{NFN}{|A|}
\end{equation}
respectively. Pairwise vertical mouldings denote size of the contained region. As well known, values of FPR and FNR are both in $[0,1]$ with a smaller value indicating a better match between the segmentation and the ground truth. On the other side, the definition DSC can be written as
\begin{equation}
DSC = \frac{2|A \cap B|}{|A|+|B|}
\end{equation}
where $\cap$ is the intersection operator. Values of DSC are in the interval of $[0,1]$ with a higher value indicating a better match between the segmentation result B and the ground truth A.

Quantitative comparison of segmentation results of the proposed model with state-of-the-art models on the BrainWeb and IBSR images in terms of FPR, FNR, and DSC are given in Fig.~\ref{fig:bwmetriccmp} and Fig.~\ref{fig:ibsrmetriccmp}, respectively. For the BrainWeb dataset, it can be seen that boxes of WM, GM, and CSF of the proposed model in terms of FPR and FNR are much more compacted and the mediums are lower than CV, LIC, and LINC which indicates segmentation results of the proposed model match better with corresponding ground truth than state-of-the-art models. On the contrary, boxes of WM, GM, and CSF of the proposed model in terms of DSC are also compacted besides the mediums are higher than CV, LIC, and LINC, which indicates better match of the segmentation results with corresponding ground truth. On the other side, for the IBSR dataset, FPR boxes of WM and FNR boxes of GM and CSF are more compacted and lower than state-of-the-art models. DSC boxes of WM, GM, and CSF are more compacted than other models with medium values similar to CV but higher than LIC and LINC. As shown in Fig.~\ref{fig:ibsrimgcmp}, biases of IBSR images are weak than BrainWeb and ground truths in IBSR images consider more non-zero area as gray matter and therefore decrease areas of WM and CSF. This is the main reason that performance of the proposed model on IBSR is worse than that on BrainWeb images. It has to be pointed out that we set $\lambda_1=2.0$ to suppress the areas considered as WM by the proposed model and impact of weighting coefficients will be discussed in Section \ref{sec_impwc}.

{{Due to potential relatedness of the proposed model to MICO as mentioned earlier, it is necessary to compare them quantitatively on public image repositories, beside giving theoretical differences in Remarks 8-10 and comparing them qualitatively on selected image slices of public datasets in section \ref{sec:qualcomp}. As shown in Fig.~\ref{fig:bwmetriccmp}, box compactnesses of the proposed model and MICO are similar. In addition, although MICO achieves higher DSC and lower FPR and FNR than the proposed model, there are obviously outliers for MICO, which is due to its sensitive to noise as mentioned earlier. For IBSR images, the proposed model achieves higher and more compact DSC boxes than MICO. This is because there are not so much strong bias in IBSR images but MICO prefers to provide higher bias estimations at image centres than the proposed model. Moreover, the proposed model achieves better FPR for WM and CSF and better FNR for GM than MICO.}}

\begin{figure*}[htp]
\centering
{\includegraphics[width=0.75\textwidth]{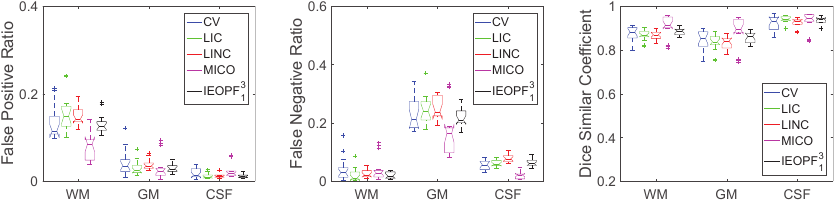}}
\caption{Quantitative comparison with state-of-the-art methods on BrainWeb images.\label{fig:bwmetriccmp}}
\end{figure*}

\begin{figure*}[htp]
\centering
{\includegraphics[width=0.75\textwidth]{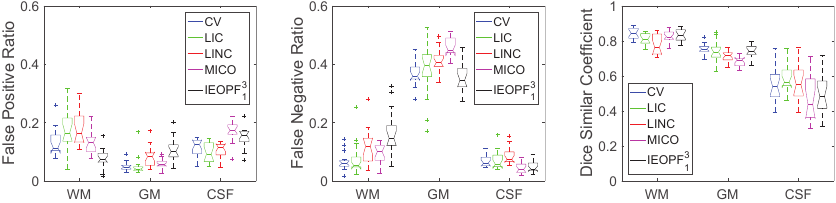}}
\caption{Quantitative comparison with state-of-the-art methods on IBSR images.\label{fig:ibsrmetriccmp}}
\end{figure*}

\section{Discussions}
\label{sec_discuss}
\subsection{Relationship with CV and PS}
\label{sec_relatewother}
It is worth pointing out that the proposed energy ${\cal{F}}$ in Eq.~(\ref{eq_img_term}) reduces to the first term of Eq.~(\ref{eq_CV_dataterm}) when 1) $\lambda_1=\lambda_2=1.0$, and 2) $w_1=1.0$ and $w_i=0$ for $i=2,3,...,M$ which indicates $b({\bf x})={\bf 1}$. That is to say the proposed model $IEOPF^2$ defined in Eq.~(\ref{eq_img_term}) is a generalization of the well known Chan-Vese model. If we define $u_i({\bf x})={\bf w}^T G({\bf x})c_i$, the energy defined in Eq.~(\ref{eq_img_term}) will reduce to the first term of Eq.~(\ref{eq_PS_dataterm}) and the smoothness of $u_i({\bf x})$ are ensured by the orthogonal primary functions $g_1$, $g_2$, ..., and $g_M$ implied in $G$. Therefore, no further regularization term like the second term in Eq.~(\ref{eq_PS_dataterm}) are needed to smooth $u_i({\bf x})$.
\subsection{Improvement to LINC}
\label{sec:cmp2LINC}
As described in \citep{feng2016image_LINC}, in the case of two phase implementation of the LINC model, there are 7 convolutions in the size of normalized kernel $K$ for each iteration of the level set function, which are the main factor causing computational burden of LINC. As smoothness of bias fields existing in images with inhomogeneous intensities can be guaranteed by orthogonal primary functions, the proposed model IEOPF removes the convolution kernel $K$ from the LINC model and therefore there is no convolution in iterations of the level set function any more.

\subsection{Robustness of IEOPF to Initialization}
\label{sec:initrobust}
As mentioned above, the proposed model is a generalization of CV and a simplification of LINC. It is well known that the intensity constants in CV can be seen as global average of inside and outside regions separated by the $0$-level set contour. Therefore, CV is greatly  non-sensitive to local intensities and robust to initialization \citep{chan2001active}. On the other side, as pointed out in \citep{feng2016image_LINC}, LINC is also robust to initialization. Thus, as a generalization of CV and a simplification of LINC, the proposed model is robust to initialization. We give a demonstration of the proposed model on one vessel image in four initialization strategies in Fig.~\ref{fig:robust2init} to verify robustness of the proposed model to initialization. It is obvious that there are not obvious differences between any two strategies in terms of bias estimation and final segmentations, which proves that the proposed model is robust to initialization.

\begin{figure}[!h]
\centering
{\includegraphics[width=0.75\textwidth]{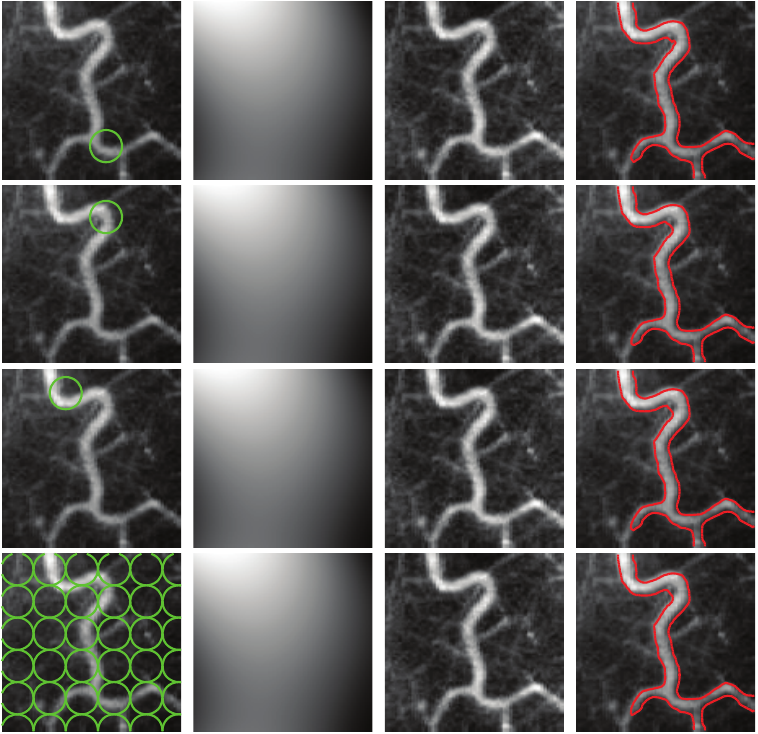}}
\\
\scriptsize\leftline{\qquad\qquad\qquad\qquad\qquad\quad Init \qquad\qquad\qquad\qquad\quad Bias \qquad\qquad\qquad\qquad Corrected \qquad\qquad\qquad\qquad Seg }
\caption{Demonstration of robustness to initialization of the proposed model.\label{fig:robust2init}}
\end{figure}

\subsection{Impact of weighting coefficients}
\label{sec_impwc}
For three phase segmentation of the proposed model on BrainWeb and IBSR datasets where $M_1=(1-H(\phi_1))(1-H(\phi_2))$, $M_2=(1-H(\phi_1))H(\phi_2)$, and $M_3=H(\phi_1)$, the formulation in Eq.~(\ref{eq_iter_phi}) can be rewritten into
\begin{eqnarray}
\label{eq_iter_phi1}
\frac{\partial\phi_1}{\partial t}&=&-\delta(\phi_1)(-\lambda_1 e_1 (1-H(\phi_2)) -\lambda_2 e_2 H(\phi_2) + \lambda_3 e_3 ) \nonumber \\
&+&\mu \left(\triangledown^2\phi_1 - {{\rm div}}\left(\frac{\triangledown\phi_1}{\mid\triangledown\phi_1\mid}\right)\right)
+\nu \delta(\phi_1){{\rm div}}\left(\frac{\triangledown\phi_1}{\mid\triangledown\phi_1\mid}\right)
\end{eqnarray}
and
\begin{eqnarray}
\label{eq_iter_phi2}
\frac{\partial\phi_2}{\partial t}&=&-\delta(\phi_2)(-\lambda_1 e_1 (1-H(\phi_1)) +\lambda_2 e_2 (1-H(\phi_1)) )  \nonumber \\
&+&\mu \left(\triangledown^2\phi_2 - {{\rm div}}\left(\frac{\triangledown\phi_2}{\mid\triangledown\phi_2\mid}\right)\right)
+\nu \delta(\phi_2){{\rm div}}\left(\frac{\triangledown\phi_2}{\mid\triangledown\phi_2\mid}\right).
\end{eqnarray}
It is obvious that $e_i({\bf x}) \geq 0$ in Eq.~(\ref{eq_iter_phi}) and $M_i\in [0,1]$ where $i=1,2,3$. Therefore, the first term on the right hand of Eq.~(\ref{eq_iter_phi1}) is monotone increasing for $\lambda_1$ and $\lambda_2$ and decreasing for $\lambda_3$ respectively, only if they take positive values. Thus, given a positive increment on $\lambda_1$ and $\lambda_2$, the level set function $\phi_1$ will be increased much harder in each iteration. On the contrary, given a positive increment on $\lambda_3$, $\phi_1$ will be decreased much harder. As described in {Algorithm \bf{\ref{alg_BCELS}}}, we let the level set functions take negative and positive values inside and outside the $0$-level set contours, respectively. Hence,for all the others fixed, the greater the coefficient $\lambda_1$ and $\lambda_2$ are, the smaller the region inside the $0$-level set contour is, and vice versa. Similarly, the greater the coefficient $\lambda_3$ is, the
smaller the region outside the $0$-level set contour is,and vice versa. Same analysis can be applied to Eq.~(\ref{eq_iter_phi2}) to conclude that the greater the coefficient $\lambda_1$ and $\lambda_2$ are, the smaller the regions inside and outside the $0$-level set contour are, and vice versa. As mentioned earlier, the regularization term and arc length term are used to maintain regularity of the level set function and smooth $0$-level set contour. Thus, the greater the parameters $\mu$ and $\nu$ are, the level set function is more close to sign distance function and the smoother the $0$-level set contour is.
 
\section{Conclusion and future work}
\label{sec_con_future}
The proposed model is effective in segmenting images with inhomogeneous intensities and provide a smooth bias estimation of the inhomogeneity. We will further improve the proposed model to extract brain tissues in 3D on public image repositories in our future work.

\section*{Acknowledgments}
This work was supported by the National Natural Science Foundation of China under grants 61602101 and
U1708261, the Fundamental Research Funds for the Central Universities of China under grant N161604003,
and the National Key Research and Development Program of China under grant 2017YFB1400804.


\end{document}